\documentclass[11pt]{article}
\usepackage[linesnumbered,boxed]{algorithm2e}
\usepackage{amssymb,amsfonts,amsmath,amsthm,cite}
\usepackage{graphicx,epsfig,subfigure}
\usepackage{enumerate}
\usepackage{tabularx}
\usepackage{dsfont}
\usepackage{booktabs,dcolumn}
\usepackage{color}
\usepackage{multirow}
\usepackage{spconf}
\usepackage{}

\def \ff{{\textbf{\emph{f}}}}
\def \gg{{\textbf{\emph{g}}}}

\def \ww{{\textbf{\emph{w}}}}
\def \xx{{\textbf{\emph{x}}}}
\def \yy{{\textbf{\emph{y}}}}
\def \xxx{{\textbf{{x}}}}
\def \yyy{{\textbf{{y}}}}
\def \uuu{{\textbf{{u}}}}
\def \vvv{{\textbf{{v}}}}
\def \zz{{\textbf{\emph{z}}}}

\newtheorem{defn}{Definition}[section]
\title{Crowd Counting Considering Network Flow Constraints in Videos}

\name{Liqing Gao$^{\star}$ \qquad Yanzhang Wang$^{\star}$ \qquad Xin Ye$^{\star}$ \qquad Jian Wang$^{\dagger}$}

\address{$^{\star}$ Institute of Information and Decision Technology \\
Dalian University of Technology, \\
Dalian 116023, P.R. China \\[10pt]
$^{\dagger}$ Department of Mathematics \\
Taiyuan University of Technology, \\
Taiyuan 030024, P.R. China}

\begin{document}
\maketitle
\begin{abstract}
The growth of the number of people in the monitoring scene may increase the probability of security threat,
 which makes crowd counting more and more important. Most of the existing approaches estimate the number of pedestrians within one frame, which results in inconsistent predictions in terms of time. This paper, for the first time, introduces a quadratic programming model with the network flow constraints to improve the accuracy of crowd counting.
 Firstly, the foreground of each frame is segmented  into groups, each of which contains several pedestrians. Then, a regression-based map is developed in accordance with the relationship between low-level features of each group and the number of people in it. Secondly, a directed graph is constructed to simulate constraints on people's flow, whose vertices represent groups of each frame and arcs represent people moving from one group to another. Then, the people flow can be viewed as an integer flow in the constructed digraph. Finally, by solving a quadratic programming problem with network flow constraints in the directed graph, we obtain consistency in people counting. The experimental results show that the proposed method can reduce the crowd counting errors and improve the accuracy. Moreover, this method can also be applied to any ultramodern group-based regression counting approach to get improvements.
\end{abstract}
\section{Introduction}\label{sec-intro}
Crowd counting is one of the most important tasks for intelligent video surveillance systems. It has a wide range of applications, such as public security, public transportation monitoring etc. Crowd gathering often happens in the monitoring scenarios, so accurately calculating and controlling the number of people can effectively reduce the probability of the abnormal events. However, crowd counting is a challenging task due to the heavy long-term occlusion and various perspective-related distortions in different surveillance environments.

Nowadays, most of the existing approaches of people counting estimate the number of people within one frame, which may lead to different counting results estimated for the same group of people in different frames. %and different counting results may be estimated in the same group of people in different frames by these methods.
%which may result in inconsistent counting results. For an example, the same group of people may be assigned different counting results in different frames., that is inconsistent counting results, which will further influence the accuracy of the results The reason
One situation is that the foreground size of the same group of people may be very large in the first frame, and then it becomes very small due to occlusion or perspective distortions in the next frame. Accordingly, a method which considers foreground size as the main feature may lead to different results for this group of people. Clearly, it influences the accuracy of the results. It cannot guarantee the global consistency of the counting results of same group of people among frames, although contemporary clues have been used in some image feature designs. Motivated by recent approaches based on graph theory for multi-object tracking tasks, the paper proposes a digraph model to represent relationship of different groups of moving people among frames in videos. Further, based on the counting results for each group in each frame as output from regression-based counting algorithms, a quadratic programming method is proposed, which is characterized with network flow constraints, to improve these counting results.

\begin{figure}[bht]
  \centering
  \centerline{\includegraphics[scale=0.5]{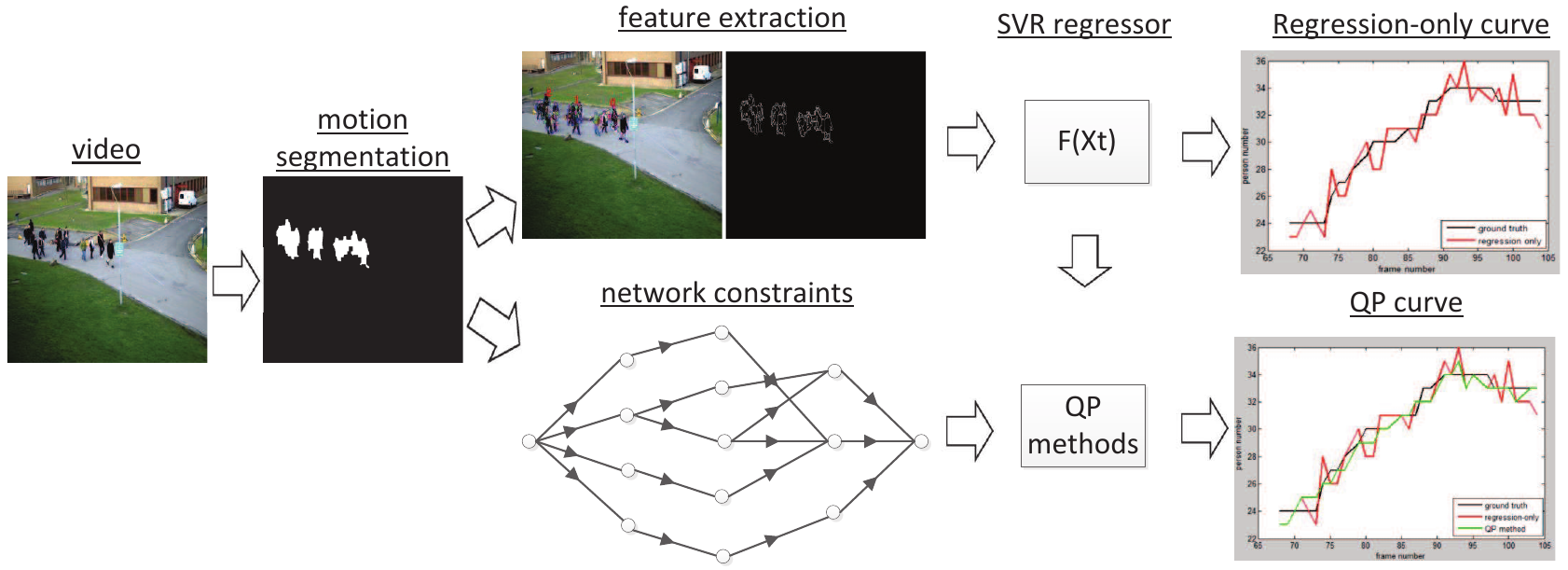}}
\caption{The main components of our method}
\label{fig-diagram}
\end{figure}

The framework of the proposed method is illustrated in Fig.\ref{fig-diagram}. It is assumed that all the moving objects are pedestrians. Firstly, the pedestrian foreground of each frame is segmented and clustered into groups. Here, we simply consider each connected region of foreground as a group. If several people are occluded with each other, then they are categorized into the same group. After extracting the features of each group, a trained support vector regressor is used to obtain the corresponding number of people.
 Further, a network is constructed with the network flow constraints, defining that the number of people entering the group equals the number of people  exiting, in which the foreground groups identified with vertices and relationships between the two groups with arcs. Then the predicted
 number of people of each group is improved by solving a quadratic programming model with network flow constraints. Finally, the counting results of each frame are obtained through totalling all groups in the same frame. The proposed quadratic  programming method can improve the performance of the regression-based counting approaches which need segmenting foreground as the first step, such as the methods previously reported \cite{CABL3,ZLN,AZZQH,ZXH,CDF2}.

Experimental results on benchmark datasets show that the proposed quadratic programming method following regression model can reduce the counting errors and obtain higher accuracy than the single regression model. Further some experiments are conducted to compare the results of the proposed algorithm with those of other advanced methods on benchmark datasets, which verified a better performance in most cases.

The paper is organized as follows: Section \ref{sec-rw} gives the  reviews of related work. Section \ref{sec-regression}
introduces the simple features and SVR regression method used for experiments in this paper. In section
\ref{sec-constraint}, we develop the network flow
model and introduce the constraints for crowd counting.
 Section \ref{sec-MPL} presents the
quadratic programming model and the solution of this model. Experimental
results of different methods applied to the crowd counting problem are
presented in section \ref{sec-exper-res}. Finally, section
\ref{sec-conclusion} gives some concluding remarks.
\section{Related Work}\label{sec-rw}
In the past decades, there were mainly three types of counting techniques\cite{RDD,SSAMS}:
%,  and counting by tracking

 \textbf{Counting by detection:} This kind of method allows to count people by a detector designed to detect each individual, for example, pedestrian detector\cite{KRSM}, face detector and head-shoulder detector\cite{LMZ,GCL}. In pedestrian detection approach, a binary classifier is trained using common features, such as Haar wavelets and  histograms of oriented gradients (HOG)\cite{HOG}. Then the trained classifier can be applied  to search for pedestrians by sliding window in the image pyramid. The detection performance can be further improved by deformable parts model\cite{DPM}. Pedestrian detection is distortion insensitive due to pyramid window search and deformable parts model, which leads to cross-scene counting techniques. However, despite the remarkable improvement, the accuracy of counting by detection seriously suffers from high missing rate of detectors, especially in high occlusion level.

\textbf{Counting by statistics}: These methods adopt machine learning techniques directly to learn a mapping from low-level features to people counting in a scene.
 Among extensive machine learning methods, regression methods\cite{CABL1,CABL2,CABL3,CDF1,CDF2,CKG,LCC,ZZW,ZLN,AZZQH} are the most popular in crowd counting. Chan and Vasconcelos \cite{CABL1,CABL2,CABL3} utilized Gaussian process regression method and Bayesian-Poisson regression methods to obtain the correspondence between the features
 of each segmented region and crowd number. D.Conte et al \cite{CDF1,CDF2} applied  support vector regression method to learn the mapping from features based on the salient points to crowd number. Al-Zaydi et al \cite{AZZQH} proposed a piece-wise linear model with dynamic features selection to deal with low and high occlusions. Dan Kong et al\cite{KGT} introduced a viewpoint invariant feature and used a single hidden layer neural network as a regressor. Yang Cong et al \cite{CGZ} estimated the number of
pedestrians by quadratic regression, and a novel feature based on flow velocity field estimation was considered as input of quadratic regression. Jingwen Li\cite{LHL} removed non-pedestrian moving objects first by template matching method, followed by a linear regressor trained to predict the number of people.
Although these type methods need some elaborate work, including feature selection and off-line training stage, they are more robust and efficient than that of detection based for a high-density crowd scene. Therefore, they gain extensive popularity in crowd counting problem. There are other machine learning methods applied to crowd counting, such as sparse respresentation\cite{FHR} and deep learning\cite{ZCL}. However, almost all statistic methods estimate the
number of people within one frame, which may result in inconsistent predictions. It means that these methods may get different values for the same group of people in consecutive frames.

\textbf{Counting by tracking:}
Recent tracking methods consider people flow as a network flow and each individual's walking path as a continuous trajectory, which can be modeled as a $1$-flow (or a path) in the network. Then it's possible to utilize network flow methods to complete tasks connected with tracking multiple objects. Anton Milan et al\cite{MAL} modeled the problem as a multi-labeled conditional random field on the network, and found a set of continuous trajectory  by $\alpha$-expansion algorithm. Horesh Ben Shitrit et al\cite{BSHB} formulated the problem of tracking multiple people whose walking paths might intersect, as a multi-commodity network flow problem. As multiple objects tracking tasks need to identify every single person, time-independent people detectors are used to detect possible locations of individuals per frame, then these locations are linked into consistent trajectories by global optimization on networks. Thus, this method may be successful in the situation where pedestrians are well-separated from each other in the frame. However, it is hard to deal with ID switches and track each person accurately in a crowd. It should be pointed out that the temporal cue and the network model are also widely used in feature selection\cite{HYW,HCC} and other applications besides multi-object tracking.

Motivated by the network flow methods in multi-object tracking, the research proposes a quadratic programming method with network flow constraints for the refinement of the counting results from regression method. Viewed as a whole object to be tracked, each group in the foreground is identified as a vertex of a network (or a directed graph). Consistent counting results, which satisfy the network flow constraints on the network, can be obtained by solving a quadratic programming problem.

\section{Basis of Regression based Counting Methods}
\label{sec-regression}

%\subsection{Foreground Segmentation}

These methods have two hypotheses (which, anyway, are those used by most of the regression based approaches in the literature): The camera is stationary; the moving objects are all pedestrians. Counting by regression algorithms usually consists of three steps: (a) segmenting the foreground to several groups; (b) extracting efficient features from the foreground groups, and (c) using a regression model to predict the number of individuals for the extracted features obtained from each foreground group.

Inter-frame difference method and other methods are used to obtain motion foregrounds, and the erosion
and dilation operations can further reduce the noise of foreground images. Then each connected region in the foreground image is
viewed as a pedestrian group. As shown in Fig.\ref{fig-3layer}(a), the
pedestrians are clustered into four groups.

%e.g. group features

Let $P$ be one of the groups in the segmented foreground image, $V$ be the feature vector $(x_c,y_c,w,h,\psi,\ell,\zeta,\theta)$, which is extracted from group $P$, let $f$ be a regression function trained by the SVR with the key-values (group feature vector,count number). So the count number $n_{P}$ of group $P$ can be estimated by
\begin{equation}
n_{P}=f(V)
\end{equation}
 where
\begin{itemize}
\item $(x_c,y_c)$ is the center of gravity of group $P$.
\item $(w,h)$ are the width and height of bounding rectangle of group $P$.
\item $\psi$ is the total number of pixels of group $P$.
\item $\ell$ is the perimeter of group $P$.
\item $\zeta$ is the total number of edge pixels contained in group $P$.
\item $\theta$ is the total number of SURF feature points contained in group $P$.
\end{itemize}
 %We convert $C$ into a feature vector, then feed it into a regressor. The output of the regressor is the estimated number of persons in the group.
%We utilize support vector regression algorithm to train a regressor. For testing, we use the trained regressor to estimate the number of individuals in each group.
\section{Network Flow Constraints for Crowd Counting and Property}
\label{sec-constraint}
\begin{figure*}[htb]
\begin{minipage}[b]{0.3\linewidth}
  \centering
  \centerline{\includegraphics[scale=0.4,bb=0 0 310 130]{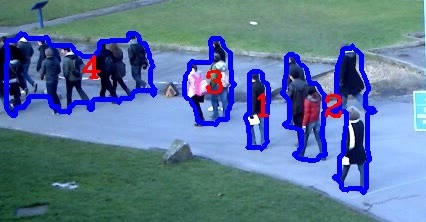}}
%  \vspace{1.5cm}
  \centerline{(a) Frame 1}\medskip
\end{minipage}
\hfill
\begin{minipage}[b]{0.3\linewidth}
  \centering
  \centerline{\includegraphics[scale=0.4,bb=0 0 310 130]{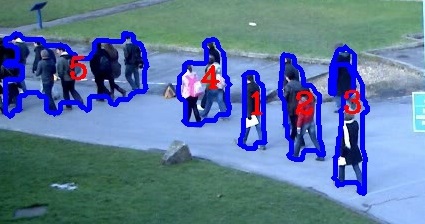}}
  \centerline{(b) Frame 2}\medskip
\end{minipage}
\hfill
\begin{minipage}[b]{0.3\linewidth}
  \centering
  \centerline{\includegraphics[scale=0.4,bb=0 0 310 130]{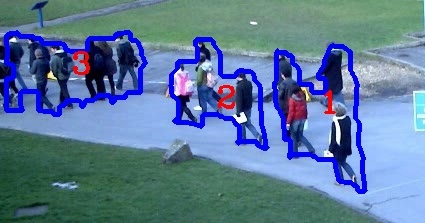}}
%  \vspace{1.5cm}
  \centerline{(c) Frame 3}\medskip
\end{minipage}
\caption{Three consecutive frames and their segmentations.}
\label{fig-3layer}
\end{figure*}
\begin{figure*}[htb]
\begin{minipage}[b]{0.3\linewidth}
  \centering
  \centerline{\includegraphics[scale=0.5,bb=0 0 210 150]{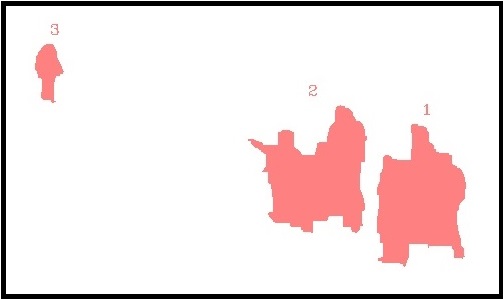}}
%  \vspace{1.5cm}
  \centerline{(a) Frame 1}\medskip
\end{minipage}
\hfill
\begin{minipage}[b]{0.3\linewidth}
  \centering
  \centerline{\includegraphics[scale=0.5,bb=0 0 210 150]{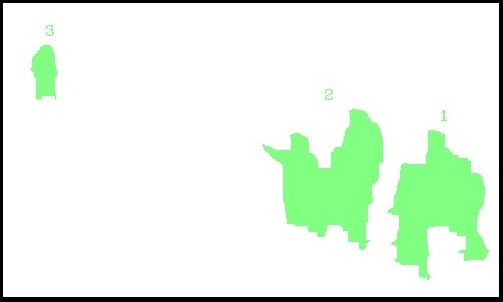}}
  \centerline{(b) Frame 2}\medskip
\end{minipage}
\hfill
\begin{minipage}[b]{0.3\linewidth}
  \centering
  \centerline{\includegraphics[scale=0.5,bb=0 0 210 150]{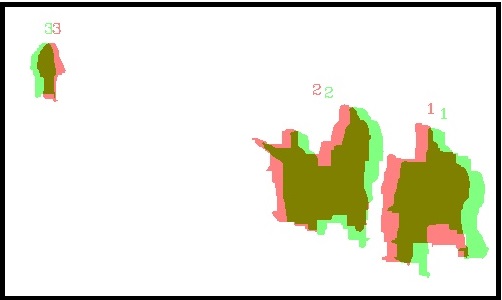}}
%  \vspace{1.5cm}
  \centerline{(c) Overlaps}\medskip
\end{minipage}
\caption{Groups of two consecutive frames and their overlaps.}
\label{fig-2layer}
\end{figure*}

Let $(I_1,I_2,\ldots,I_n)$ be a sequence of frames in a video. For each frame $I_i$, the foreground of $I_i$ is segmented into $m_i$ groups $P_1^i,P_2^i,\ldots,P_{m_i}^i$ and each group is an integral region in frame $i$. Let $P^i=\{P_1^i,P_2^i,\ldots,P_{m_i}^i\}$ and $P=\bigcup_{t=1}^{n}P^t$. As the temporal cue is very important for video analysis, the tracking of each group can get more information about the number of people. For example, as shown in Fig.\ref{fig-2layer}(a), the foreground of Frame 1 is segmented into three groups,  which are $P_1^1,P_2^1,P_3^1$.  Fig.\ref{fig-2layer}(b) shows three groups in Frame 2,  which are  $P_1^2,P_2^2,P_3^2$. Since the camera is fixed and people cannot move fast within two frames, thus, $P_1^1$ and $P_1^2$ have large overlap as shown in Fig.\ref{fig-2layer}(c), so do $P_2^1$ and $P_2^2$, $P_3^1$ and $P_3^2$. It is easy to see that $P_i^1$ and $P_i^2$ are the same group, they have the same number of people ($i=1,2,3$).

Therefore, for two groups in consecutive frames, it depends on the area of overlap to link them.
Moreover, two or more groups in the same frame may merge into one group in the next frame, with the number of people of merged group equalling to the total sum people of each group in the previous one. And one group may be divided into two or more groups in the next frame, the number of people in divided group equals the total sum of number of people in each group in the next one. In order to model all these kinds of situations, a digraph model is utilized to track each group. If one group in the first frame overlaps with another group in the second one, there is an arc between the two groups. The formal definition will be given in Def \ref{def-network}.

\begin{figure}[hb]
\begin{minipage}[b]{0.48\linewidth}
  \centering
  \centerline{\includegraphics[scale=0.2,bb=0 0 800 600]{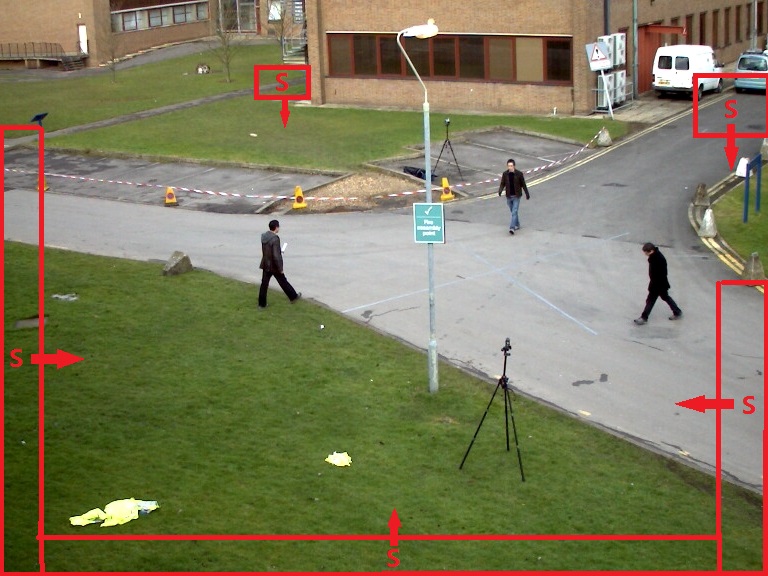}}
%  \vspace{2.0cm}
\end{minipage}
\hfill
\begin{minipage}[b]{0.48\linewidth}
  \centering
  \centerline{\includegraphics[scale=0.2,bb=0 0 800 600]{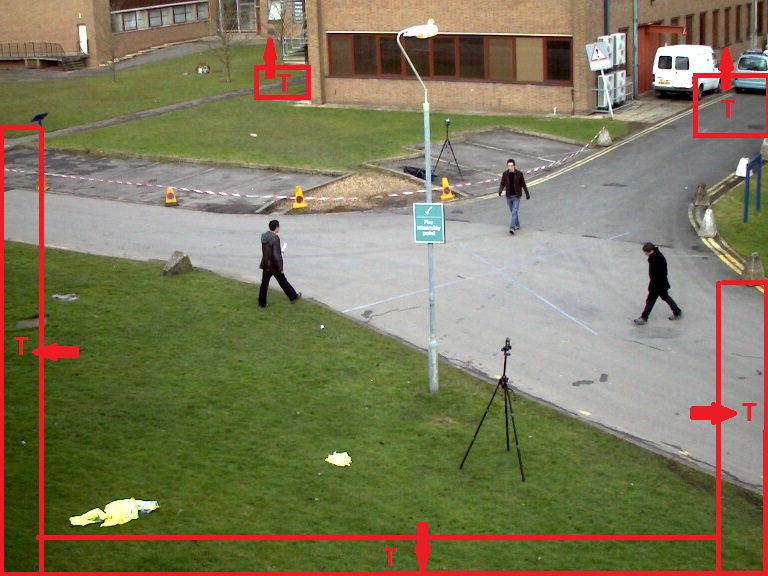}}
  %  \vspace{2.0cm}
\end{minipage}
\caption{Entering region S and exiting region T in the specific scene.}
\label{st-region}
\end{figure}

This paper denotes with S where people enter the fixed scene and with T where people exit the fixed scene. The number of people in groups entering the S-region may increase at any time, and that of groups exiting the T-region may decrease at any time. For a fixed scene, regions where people enter and exit, often close to the image boundaries, are not changed. The selection of S and T is done manually. An example is shown in Fig.\ref{st-region}. S and T are the same region, because they both allow pedestrian to enter and exit in these videos.

\begin{defn}\label{def-network}
Let $D(V,A)$ be a directed graph, whose vertex set is $V=P\bigcup\{S,T\}$ and the arc set is $A$ , as in: \\
\begin{itemize}
\item[] (1) $\langle P_i^t,P_j^{t+1}\rangle \in A$ if and only if  $P_i^t$ overlaps $P_j^{t+1}$;\\
\item[] (2) $\langle S,P_i^t\rangle\in A$ if $P_i^t$ overlaps $S$, and $\langle P_i^t,T\rangle\in A$ if $P_i^t$ overlaps $T$;\\
\item[] (3) $\langle S,P_i^t\rangle\in A$ if frame $t$ is the first frame of image sequence, and $\langle P_i^t,T\rangle\in A$ if frame $t$ is the last frame of image sequence;\\
\item[] (4) $\langle S,P_i^t\rangle\in A$ if $P_i^t\cap P_k^{t-1} = \emptyset $ for any $P_k^{t-1}\in P^{t-1}$, and $\langle P_i^t,T\rangle\in A$ if $P_i^t\cap P_j^{t+1} = \emptyset $ for any $P_j^{t+1}\in P^{t+1}$.
\end{itemize}
\end{defn}

Clearly, directed graph $D$ is a network with regions $S$ and $T$. It should be noticed that all groups in the first frame have arcs pointed from $S$ and all groups in the last frame have arcs pointed to $T$  according to Def \ref{def-network}(4). Since frame number is increasing with the extension of each arc, then network $D$ is acyclic. To put it simple, groups and vertices of graph $D$ are viewed as equivalent in this paper. Obviously, foreground segmentation may not be as accurate as expected. Group tracking failures may be caused by the fact that some $P_i^t$ has no intersection with any groups in $P^{t-1}$ or some $P_i^t$ has no intersection with any groups in $P^{t+1}$. If $P_i^t$ has no intersection with any groups in $P^{t-1}$, then  $\langle S, P_i^t \rangle$ is an arc according to Def \ref{def-network}(4). If $P_i^t$ has no intersection with any groups in $P^{t+1}$, then $\langle P_i^t, T \rangle$ is an arc according to Def \ref{def-network}(4). The robustness of the proposed method will be shown as follows.

\begin{figure}[bht]
  \centering
  \centerline{\includegraphics[width=7cm]{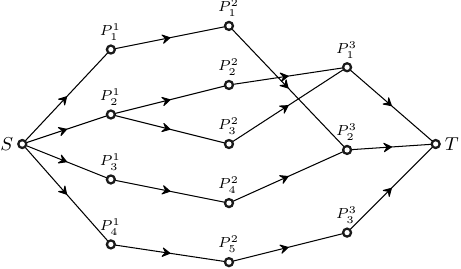}}
\caption{3-layer network $H(2,1)$ corresponding to three frames in Fig.\ref{fig-3layer}.}
\label{fig-3layernetwork}
%\textcolor{red}{Every vertex (i,j) should be written as $P^{i}_{j}$}
\end{figure}

%\subsection{\textcolor{red}{??Definition of} Finite Layer Network}
Since it is impossible to construct the network of whole video in real time, for any frame $I_c$, the set of $(2\ell+1)$ frames $\{I_{c-\ell},I_{c-\ell+1},\ldots, I_{c-1},I_c,I_{c+1},\ldots,I_{c+\ell}\}$ is used to construct the network. The graph is called $(2\ell+1)$-layer network, which is centered in frame $c$ and denoted by $H(c,\ell)$. The three consecutive frames are shown in Fig.\ref{fig-3layer}, and the corresponding 3-layer network centered in frame $2$, denoted by $H(2,1)$, is shown in Fig.\ref{fig-3layernetwork}.

%Tuple $(i,j)$ in the figure represents the $j$-th group in $i$-th frame.

\begin{figure}[htb]
\center
\includegraphics[scale=0.9]{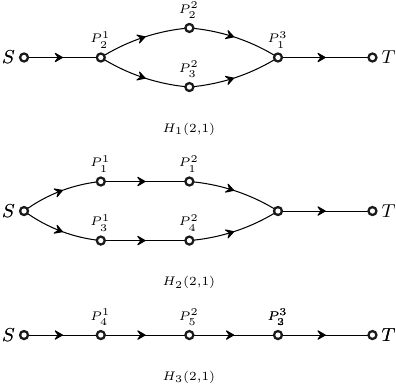}
\caption{Connected component decomposition of network $H(2,1)$.}
\label{ccd}
%Connected component decomposition of Networks Distribution {\color{blue}connecting} components in the network
\end{figure}

For digraph $D$, the sub-graph induced by $V_1\subset V$ is
denoted by $D[V_1]$. Let $D'= D[V\backslash \{S,T\}]$, and decompose $D'$ into $p$ weakly connected components $D'_1, D'_2, \ldots, D'_p$. Then for each $i\in \{1,2,\ldots,p\}$, induced digraph $D[V(D'_i)\cup \{S,T\}]$ forms a network with region $S$ and $T$ , which is called weakly connected sub-networks $D_i$. As an example, the three weakly connected sub-networks corresponding to network $H(2,1)$ in Fig.\ref{fig-3layernetwork} are shown in Fig.\ref{ccd}.

%\subsection{Network Flow Constraints}
Suppose that the segmentation of foreground never divides a single person into two or more groups, then the actual number of people in each group is an integer value. The definition of function $\ff$ is as follow. %we may define $\ff$ as a function
\begin{equation}
\ff: A \mapsto \mathds{N}.
\end{equation}
such that
\begin{itemize}
\item For any arc $\langle P_i^t, P_j^{t+1} \rangle\in A$, define $\ff(\langle P_i^t, P_j^{t+1} \rangle)$ as the actual number of people moving from group $i$ in frame $t$ to group $j$ in frame $t+1$;
\item For any arc $\langle S, P_i^t\rangle\in A$, define $\ff(\langle S, P_i^t \rangle)$ as the actual people counting of group $i$ in frame $t$;
\item For any arc $\langle P_i^t,T \rangle\in A$, define $\ff(\langle  P_i^t,T \rangle)$ as the actual people counting of group $i$ in frame $t$.
\end{itemize}

 Let
 \begin{equation}
\begin{array}{ll}
\ff^+(P_i^t) &= \sum\limits_{\langle P_i^t,\yy \rangle\in A}{\ff(\langle P_i^t,\yy \rangle)},\\[8pt]
\ff^-(P_i^t) &= \sum\limits_{\langle \xx,P_i^t \rangle \in A} {\ff(\langle{\xx ,P_i^t}\rangle)}.
\end{array}
\end{equation}

 Clearly, $\ff^+(P_i^t) =\ff^-(P_i^t)$ and they both represent the actual number of people in group $P_i^t$, so $\ff(P_i^t) = \ff^+(P_i^t) =\ff^-(P_i^t)$.  Moreover, people only enter the scene from region $S$ and exit the scene through region $T$, which implies that $\ff^+(S) =\ff^-(T)$. Thus, function $\ff$ on arc set $A$ is an integer $(S,T)$-flow in network $D$. The flow constraints of network $D$ can be defined as follows:

\begin{equation}
\left\{
\begin{array}{ll}
& \sum\limits_{\langle \xx,P_i^t \rangle \in A} {\ff(\langle{\xx,P_i^t}\rangle)}= \sum\limits_{\langle P_i^t, \yy \rangle\in A} \ff(\langle P_i^t,\yy \rangle)=\ff(P_i^t), \ \mbox{ for $P_i^t\in V \backslash\{S,T\}$}\\[8pt]
&\sum\limits_{ \langle S,P_j^t \rangle \in A} \ff(\langle S,P_j^t \rangle)= \sum\limits_{\langle P_k^t,T \rangle\in A} \ff(\langle P_k^t,T \rangle)
\end{array}
\right.
\end{equation}

Therefore, the predicted number of people of $(2\ell+1)$ frames in a video, which is a consistent result, should satisfy the flow constraints of network $D$.
%when we get the predicted number of pedestrians of $(2\ell+1)$ frames in a video, a consistent result should satisfy the flow constraints of network $D$.

\section{Quadratic Programming Model}%Quadratic
\label{sec-MPL}

Let $\hat{\ff}(P_i^t)$ be the predicted value of group $P_i^t$, which is obtained by the regression method described in section \ref{sec-regression}. %The regression method described in section \ref{sec-regression} can predict the number of people for each group in the network. Let $\hat{\ff}(P_i^t)$ be the predicted value of group $P_i^t$.
Since the prediction is processed per frame, it can never guarantee consistency results, which means it will violate the network flow constraints of network $D$. In order to obtain better crowd counting results, an integer quadratic programming model on network $D$ is put forward to improve the performance of regression based methods.
 %we put forward a quadratic programming model on network $D$ as follows.
 \subsection{Quadratic Programming Model}\label{sec-QPL}
 The quadratic programming model on network $D$ is constructed as follows:
\begin{align}
min & \quad  \sum_{P_i^t \in V \backslash \{S,T\}} \ww(\hat{\ff}(P_i^t))\cdot (\ff(P_i^t)-\hat{\ff}(P_i^t))^2\nonumber\\
s.t.&\qquad \ff(P_i^t)=\sum\limits_{\langle \xx,P_i^t \rangle \in A} {\ff(\langle{\xx,P_i^t}\rangle)}, \ \mbox{ for $P_i^t\in V \backslash\{S,T\}$}\nonumber\\[8pt]
&\qquad \ff(P_i^t)=\sum\limits_{\langle P_i^t,\yy \rangle\in A} \ff(\langle P_i^t,\yy\rangle), \ \mbox{ for $P_i^t\in V \backslash\{S,T\}$}\nonumber\\[8pt]
&\qquad \sum\limits_{ \langle S,P_j^t \rangle \in A} \ff(\langle S,P_j^t \rangle)= \sum\limits_{\langle P_k^t,T \rangle\in A} \ff(\langle P_k^t,T \rangle)
\end{align}
where $\ff(P_i^t)$'s and $\ff(\langle P_i^t,P_j^{t+1}\rangle)$'s are integer valuables,
$\ww(\hat{\ff}(P_i^t))$ represents the reliability of prediction
$\hat{\ff}(P_i^t)$ for the people counting of group $P_i^t$. The determination of weights $\ww(\hat{\ff}(P_i^t))$ will be given in section \ref{sec-algorithm code}.
\subsubsection{Model Solution}
In general, it is difficult to obtain the solution of the proposed model directly. However, by relaxing the integer valuables $\ff(P_i^t)$, $\ff(\langle P_i^t,P_j^{t+1}\rangle)$ ($\forall{P_i^t}\in V \backslash\{S,T\}$, $\forall{P_j^{t+1}}\in V \backslash\{S,T\}$) into real valuables, the model is modified to a quadratic programming problem with only linear equation constraints. Since the objective function is
strongly convex, the problem has a unique optimal solution. The
Lagrangian function of (5) can be written as follows:
\begin{align}
L(\ff,{\lambda}^-,{\lambda}^+,\mu)=& \sum_{P_i^t \in V \backslash \{S,T\}} \ww(\hat{\ff}(P_i^t))\cdot (\ff(P_i^t)-\hat{\ff}(P_i^t))^2  \nonumber\\
&+ \sum_{P_i^t\in V \backslash \{S,T\}} \lambda_{P_i^t}^-\left(\ff(P_i^t)-\sum\limits_{\langle \xx,P_i^t \rangle \in A} {\ff(\langle{\xx,P_i^t}\rangle)}\right)  \nonumber\\
&+ \sum_{P_i^t \in V \backslash \{S,T\}} \lambda_{P_i^t}^+\left( \ff(P_i^t)-\sum_{\langle P_i^t,\yy \rangle\in A}\ff(\langle P_i^t,\yy \rangle)\right) \nonumber\\&+
\mu\left(\sum_{\langle S,P_j^t \rangle\in A} \ff(\langle S,P_j^t \rangle) -\sum_{\langle P_k^t,T \rangle\in A} \ff(\langle P_k^t, T \rangle)\right).
\end{align}

In mathematical optimization, the Karush-Kuhn-Tucker (KKT) conditions are necessary conditions of the first order. For convex programming problems, KKT conditions are also sufficient. The KKT condition of the proposed problem (5) can be derived as follows:
\begin{equation}\label{final-form}
\left\{
\begin{array}{ll}
\sum\limits_{\langle \xx,P_i^t \rangle \in A} {\ff(\langle{\xx,P_i^t}\rangle)} =\ff(P_i^t), \ \mbox{ for $P_i^t\in V \backslash\{S,T\}$};\\[12pt]
\sum\limits_{\langle P_i^t,\yy \rangle\in A} \ff(\langle P_i^t,\yy\rangle) = \ff(P_i^t), \ \mbox{ for $P_i^t\in V \backslash\{S,T\}$};\\[12pt]
\sum\limits_{ \langle S,P_j^t \rangle \in A} \ff(\langle S,P_j^t \rangle)= \sum\limits_{\langle P_k^t,T \rangle\in A} \ff(\langle P_k^t,T \rangle);\\[12pt]
\lambda_{P_j^t}^++\lambda_{P_k^{t+1}}^- =0,\ \mbox{for $\langle P_j^t,P_k^{t+1} \rangle \in A$};\\[12pt]
-\mu+\lambda_{P_j^t}^- =0,\ \mbox{for $\langle S,P_j^t \rangle \in A$};\\[12pt]
\lambda_{P_k^t}^++\mu =0,\ \mbox{for $\langle P_k^t,T \rangle\in A$};\\[12pt]
\sum\limits_{P_i^t \in V \backslash \{S,T\}} \ww(\hat{\ff}(P_i^t))\cdot (\ff(P_i^t)-\hat{\ff}(P_i^t))+\lambda_{P_i^t}^-+\lambda_{P_i^t}^+ =0,\ \mbox{for any $P_i^t \in V \backslash \{S,T\}$}.
\end{array}
\right.
\end{equation}
Suppose $|V|=n$ and $|A|=m$, the KKT condition is a linear system with $3n+m-5$ variables and $3n+m-5$ equations. Therefore, the solution of the quadratic programming problem can be obtained by solving KKT linear system.

As can be seen, network $D$ can be decomposed
into several weakly connected sub-networks $D_1, D_2, $\\$\ldots ,D_p$. Since each sub-network is
independent, then the original problem can be divided into some
simple sub-problems, which means we only need to solve  the quadratic programming model on each weakly connected sub-network $D_i$, for $i=1,2,\ldots,p$.

If some weakly connected sub-networks of $D$ have a directed path with the $n$
internal vertices like $H_3(2,1)$ shown in Fig.\ref{ccd}
($H_3(2,1)$), network flow constraints will turn to be
$\ff(P_{l_1}^1)=\ff(P_{l_2}^2)=\ldots=\ff(P_{l_n}^n)$. After setting them all be $f$, the optimal problem becomes as follow:
\begin{equation}
min  \quad  \sum_{t=1}^n \ww(\hat{\ff}(P_{l_t}^t))\cdot ( f -\hat{\ff}(P_{l_t}^t))^2.
\end{equation}
Finally, we get the consistent prediction, which is the weighted average value of each prediction.
\begin{equation}\label{eq-8}
{f}=\frac{\sum_{t=1}^n \ww(\hat{\ff}(P_{l_t}^t))\cdot \hat{\ff}(P_{l_t}^t)}{\sum_{t=1}^n \ww(\hat{\ff}(P_{l_t}^t))}.
\end{equation}

\subsubsection{Reduction of Lagrange Multipliers}
In order to simplify the KKT linear system, the constraints on Lagrange multipliers are analyzed. The equations in (10) are independent with other valuables in (7), and each of them is related to an arc of graph $D$.
\begin{equation}\label{eq-lag}
\left\{
\begin{array}{ll}
\lambda_{P_j^t}^++\lambda_{P_k^{t+1}}^- =0,\ \mbox{for $\langle P_j^t,P_k^{t+1} \rangle \in A$};\\[12pt]
-\mu+\lambda_{P_j^t}^- =0,\ \mbox{for $\langle S,P_j^t \rangle \in A$};\\[12pt]
\lambda_{P_k^t}^++\mu =0,\ \mbox{for $\langle P_k^t,T \rangle\in A$}.
\end{array}
\right.
\end{equation}

\begin{figure}[bht]
  \centering
  \centerline{\includegraphics[scale=1]{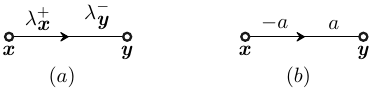}}
\caption{(I) The arc representation of Lagrange  multipliers. (II) Reduction of Lagrange  multipliers.}
\label{fig-edgereps}
\end{figure}

Let $\lambda_S^+=-\mu$ and $\lambda_T^-=\mu$, and set $\bar{A}$ be the union set of $A$ and $\{ \langle S,T\rangle\}$.  Then equations \eqref{eq-lag} can be unified as equation \eqref{eq-lagmulti}.
 \begin{equation}\label{eq-lagmulti}
\lambda_{\xx}^++\lambda_{\yy}^-=0,\ \mbox{for any $\langle \xx,\yy\rangle \in \bar{A}$}.
\end{equation}
For any arc $\langle \xx,\yy\rangle \in \bar{A}$, Lagrange multiplier $\lambda_{\xx}^+=-a$ can be viewed as out-weight of arc $\langle \xx,\yy\rangle$,
$\lambda_{\yy}^-=a$ can be viewed as in-weight of $\langle \xx,\yy\rangle$. Fig.\ref{fig-edgereps} shows an edge representation of two Lagrange  multipliers.
Define $Deg^+(\xx)=\{\yy|\langle \xx,\yy \rangle\in \bar{A}\}$ and  $Deg^-(\xx)=\{\yy|\langle \yy,\xx \rangle\in \bar{A}\}$. Clearly, equation \eqref{eq-lagmulti} implies  $\lambda_{\xx}^-=\lambda_{\yy}^-$ for any $\xx,\yy \in Deg^+(\zz)$ and $\lambda_{\xx}^+=\lambda_{\yy}^+$ for any $\xx,\yy \in Deg^-(\zz)$.
\begin{defn}
For any two arcs $\langle \uuu,\vvv \rangle \in \bar{A}$ and $\langle \xxx,\yyy \rangle \in \bar{A}$, define relation $R$ on arc set $\bar{A}$ as  $\langle \uuu,\vvv \rangle R \langle \xxx,\yyy \rangle$ if and only if $\uuu=\xxx$ or $\vvv=\yyy$.
\end{defn}

%The equivalent class of graph $D$ can be defined: $A_c$ is an equivalent class, if the arcs in $A_c$ have the same label.
According to the defined relation $R$, it is easy to
prove that the transitive closure of  $R$ is a relation of equivalence. So arc set $A$ can be partitioned into distinct $k$ equivalent classes, say $A_1, A_2,
\ldots, A_k$. For example, the equivalent classes of $\bar{A}(H_1)=A(H_1)\cup\{\langle S,T\rangle\}$ in Fig.\ref{ccd} are sets $A_1(H_1)$, $A_2(H_1)$ and $A_3(H_1)$, where
\begin{align*}
A_1(H_1) = &\{\langle S,T\rangle,\langle S,P_2^1\rangle,\langle P_1^3,T\rangle\};\\
A_2(H_1) = &\{\langle P_2^1,P_2^2\rangle,\langle P_2^1,P_3^2 \rangle \};\\
A_3(H_1) = &\{\langle P_2^2,P_1^3\rangle,\langle P_3^2,P_1^3 \rangle \}.
\end{align*}

\begin{figure}[ht]
  \centering
  \centerline{\includegraphics[scale=0.9]{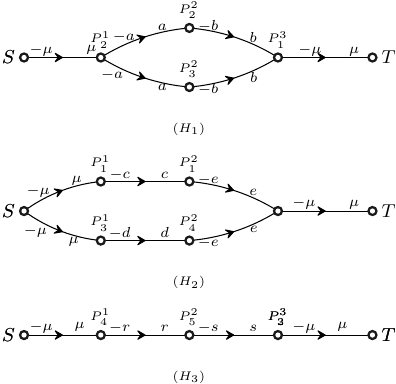}}
\caption{Equivalence classes of lagrange multipliers.}
\label{fig-eqclasses}
\end{figure}

Arcs in the same equivalent class have the same
out-weights and in-weights, and they share one common valuable.
Therefore $2|V|-4$ Lagrange multipliers are reduced to $k$
valuables. Suppose arc set $A(D)$ can be partitioned into distinct $k$ equivalent classes, say $A_1, A_2,
\ldots, A_k$. Then define $R\langle \cdot,\yy \rangle =i$ if and only if some arc $\langle \xx,\yy \rangle \in A_i$, define $R\langle \xx,\cdot \rangle =i$ if and only if some arc $\langle \xx,\yy \rangle \in A_i$. Finally by multipliers reduction, the multipliers can be reduced to $\lambda_1,\lambda_2,\ldots,\lambda_k$ and  Equation (\ref{final-form}) can be simplified to the following form.
\begin{equation}\label{final-form2}
\left\{
\begin{array}{ll}
\sum\limits_{\langle \xx,P_i^t \rangle \in A} {\ff(\langle{\xx,P_i^t}\rangle)} =\ff(P_i^t), \ \mbox{ for $P_i^t\in V \backslash\{S,T\}$};\\[12pt]
\sum\limits_{\langle P_i^t,\yy \rangle\in A} \ff(\langle P_i^t,\yy\rangle) = \ff(P_i^t), \ \mbox{ for $P_i^t\in V \backslash\{S,T\}$};\\[12pt]
\sum\limits_{ \langle S,P_j^t \rangle \in A} \ff(\langle S,P_j^t \rangle)= \sum\limits_{\langle P_k^t,T \rangle\in A} \ff(\langle P_k^t,T \rangle);\\[12pt]
\sum\limits_{P_i^t \in V \backslash \{S,T\}} \ww(\hat{\ff}(P_i^t))\cdot (\ff(P_i^t)-\hat{\ff}(P_i^t))+\lambda_{R \langle \cdot, P_i^t\rangle }+\lambda_{R \langle P_i^t,\cdot\rangle } =0,\ \mbox{for any $P_i^t \in V \backslash \{S,T\}$}.
\end{array}
\right.
\end{equation}
The arc representation of reduced Lagrange multipliers of $H_1$, $H_2$ and $H_3$ is given in Fig.\ref{fig-eqclasses}, and all the Lagrange multipliers are labeled on the arcs of the network and
arcs with the same label are equivalent. As shown in Fig.\ref{fig-eqclasses}, arcs in $A_1(H_1)$ share one common Lagrangian multiplier $\mu$. Arcs in $A_2(H_1)$ share one common Lagrangian multiplier $a$ and arcs in $A_3(H_1)$ share one common Lagrangian multiplier $b$.
\subsection{Algorithm pseudo-code and time complexity }\label{sec-algorithm code}
 % Our proposed algorithm consists of five steps: (a) segmentation of the foreground into several groups; (b) feature extraction  from the foreground groups; (c) a regression model to predict the number of individuals for the extracted features obtained from each foreground group; (d) construction of finite layered networks, and (e) solving  quadratic programming model to get final predictions.
   Algorithm \ref{alg-1} shows the pseudo-code of our proposed method.
\begin{algorithm}
\label{alg-1}
\caption{QPL$(2\ell+1)$ method.}
\KwIn{A sequence of frames $I_1,I_2,\ldots,I_n$.}
\KwOut{Number of pedestrians $f_1,f_2,\ldots,f_n$ according to each frame.}

\For{$j=1;j \leq n;j++$}
{
 \For{$t=j-2\ell;t \le j+2\ell;t++$}
 {
    \If{$t\geq 1 \&\& t\leq n$}
    {
        Segment $I_t$ into $m_t$ groups and let $P^k=\{P_1^t,P_2^t,\ldots,P_{m_t}^t\}$;

        \For{$i=1;i\leq m_t;i++$}
        {
           Utilize trained regressor to predict people counting  $\hat{\ff}(P_i^t)$ in group $P_i^t$;
        }
    }
    \Else
    {
        Let $m_t =0$ and $P^t=\emptyset$;
    }
 }

  Construct network $H(j,\ell)$ with vertex set $V(j,\ell)=\left(\bigcup_{t=j-2\ell}^{j+2\ell} P^t \right)\cup \{S,T\}$;

  Decompose network $H(j,\ell)$ into $p$ weakly connected sub-networks $H_1(j,\ell)$, $H_2(j,\ell)$, $\ldots, H_p(j,\ell)$;

  \For{$i=1;i\leq p;i++$}
  {
      \If{$H_i(j,\ell)$ is a path}
      {
          Let $f$ be the weighted average value of all $\hat{\ff}(P_i^t)$'s in $H_i(j,\ell)$ according to Equation (\ref{eq-8});

          \For{ each vertex $v$ in $H_i(j,\ell)$}
          {
              $\ff(v)=f$;
          }
      }
      \Else
      {
        Initialize a Union Set $U$ such that each arc in $H_i(j,\ell)$ is in a single class;
        \For{ each vertex $v$ in $H_i(j,\ell)$}
        {
           Union all classes of $U$ that contain in-arcs of $v$ into one class;

           Union all classes of $U$ that contain out-arcs of $v$ into one class;
        }

        Build system of linear equations $Ax=b$ according to Equation (\ref{final-form2});

        Obtain $S,T$-flow $\ff$ on $H_i(j,\ell)$ by solving linear equations;
      }

  }
   $f_j=0$;

  \For{$i=1;i\leq m_j;i++$}
  {
      $f_j=f_j+\ff(P_i^j)$;
  }
}

return $f_1,f_2,\ldots,f_n$;
\end{algorithm}

 \textbf{Determination of weights:} In order to determine $\ww(\hat{\ff}(P_i^t))$ ($\forall P_i^t \in V \backslash\{S,T\}$) in quadratic programming model, we analyzed the trained regressor in section
\ref{sec-regression}. Let $\{(P_i^t,\gg(P_i^t))\colon 1\leq i\leq m_t,t\in I\}$ be the training set, where $I$ is the set of selected frame numbers,  $m_t$ is the number of groups in the frame $t$ and $\gg(P_i^t)$ is the actual number
of people in group $P_i^t$. Let $\bar{\ff}(P_i^t)$ denote the number of
people of each group $P_i^t$ predicted by the trained regressor.
Then the training data can be represented by set
$\{(P_i^t,\gg(P_i^t),\bar{\ff}(P_i^t))\colon 1\leq i\leq m_t,t\in I\}$.  For a given
predicted value $\tau$, there must be a list of groups whose number of people are predicted as $\tau$, that is, set
$G(\tau)=\{\gg(P_i^t)\colon \bar{\ff}(P_i^t) = \tau,1\leq i\leq m_t,t\in I\}$. %We calculate the mean and variance of $G(\tau)$.
Finally, the regressor was regulated by subtracting the mean of $G(\tau)$ when the
predicted value equals $\tau$, and let weight $\ww(\tau)$  be the variance of set $G(\tau)$.

 \textbf{Time complexity of QPL$m$ algorithm} : Suppose the image size is $m\times n$, then foreground extraction, foreground segmentation and  feature extraction will cost $O(mn)$. The SVR regressor costs $O(l)$($l$ is the feature vector length)to get predictions of each group, in which $l$ is the feature vector length. For directed graph building process, each pixel in each group is labeled with its group id. If the area of the overlap between group $P_i^t$ and group $P_j^{t+1}$ is greater than a given threshold, then there is an arc from group $P_i^t$ to group $P_j^{t+1}$.
 Thus, it costs at most $O(mn)$ to create a finite layered directed graph. Finally, suppose the  graph has $|V|$ vertices and $|A|$ arcs, then the KKT condition of equation \eqref{final-form} is a linear equation with $3|V|+|A|-5$ variables and $3|V|+|A|-5$ equations. If Gaussian elimination approach is adopted to solve this model, the time complexity is $O(|V|^3+|A|^3)$. Thus, the total time complexity is $O(mn+|V|^3+|A|^3)$. Because the proposed method only considers the limited layers, such as 23 layers, $|V|$ and $|A|$ are very small.
  %If we use Gaussian elimination, then the time complexity is $O(|V|^3+|A|^3)$. However, as we only consider at most 23 layers , $|V|$ and $|A|$ are very small. Thus, the time complexity of our algorithm is  $O(mn+|V|^3+|A|^3)$.

\section{Experimental Results and Analysis}\label{sec-exper-res}
\subsection{Experiment Setup}

 Experiments are conducted on three benchmark datasets: PETS2009 dataset ,UCSD dataset and Fudan dataset. The PETS 2009 dataset [28] is organized in four sections, but our attention is mainly focused on the section S1 that was used to benchmark algorithms for the ``Person Count and Density Estimation" of PETS2009 and 2010 contests. The experimental videos involve two different views captured by using two cameras that
 contemporaneously acquired the same scene from different points of view. They are eight videos of this dataset, namely S1.L1.13-57, S1.L1.13-59, S1.L2.14-06 and S1.L3.14-17 in view 1, and  S1.L1.13-57, S1.L2.14-06, S1.L2.14-31  and S3.MF.12-43 in view 2; which can be denoted as V1, V2, V3, V4, V5 ,V6, V7, and V8 for short.
The UCSD dataset is introduced by Chan\cite{CABL1} and
contains 2000 annotated frames of the pedestrian moving along a walkway.
The Fudan dataset is proposed by Tan\cite{TBZ} and contains five
sequences of 300 frames each, 1500 frames in total. It is worth noting that the ground-truths of all datasets
are generated by manually counting people in the specified regions
in each sampled frame.

The indices used to report the performance are the Mean Absolute Error(MAE) and the Mean Relative Error(MRE) defined as follows,
{\setlength\abovedisplayskip{0.3pt}
{\setlength\belowdisplayskip{0.3pt}
\begin{equation}%\label{eq-MAE-MRE}
{MAE}=\frac{1}{N} \cdot \sum_{i=1}^N|G(i)-T(i)| ,
\end{equation}
\begin{equation}%\label{eq-MAE-MRE}
{MRE}=\frac{1}{N} \cdot \sum_{i=1}^N \frac{|G(i)-T(i)|}{T(i)}
\end{equation}
where N is the number of frames of the test video and $G(i)$ and $T(i)$ are the estimated and the true numbers of individuals in the $i$-th frame, respectively.
%\subsection{Regulation of Regressor and Determination of Weights}
\subsection{Experimental results of QPL$m$}
On these datasets, two groups of tests are performed. For simplicity, the method that uses SVR regression only is named after RO (regression-only), and the proposed method is named after QPL$m$ (quadratic programming model with m layers).

The first group of tests is carried on PETS2009 datasets, in which inter-frame difference method was applied to obtain the foreground images. The training set is constructed by manually collecting 30-40 frames from the video, and the rest frames are used for testing.
 The experimental results are reported in Table
\ref{tab-performance}, and the curves of crowd estimation with
 two methods on four videos are shown in Fig.\ref{fig-res}.
 Table \ref{tab-performance} shows
that the errors of QP method tend to descend with the increase of the numbers of network layers. As shown in Fig.\ref{fig-res}, the RO method curve of video V1 oscillates around the ground-truth curve (Fig.\ref{fig-res}(a)). Its errors
are Gaussian-like noises, which are suited to be solved by QP method.
Thus, the QP method performs well.
\begin{table}[h]
\caption{performance of RO method and QPL$m$ on PETS2009 dataset.}\label{tab-performance}
\centering
\begin{tabular}{ccccccccc}
\hline
\multirow{2}{*}{Method} &
\multicolumn{2}{c}{RO} &
\multicolumn{2}{c}{QPL7} &
\multicolumn{2}{c}{QPL15} &
\multicolumn{2}{c}{QPL23} \\
\cline{2-3}\cline{4-5}\cline{6-7}\cline{8-9}
  & MAE & MRE &  MAE & MRE & MAE & MRE & MAE & MRE \\
\hline
V1 & 1.29 & 5.58\% & 1.21 & 5.27\% & 1.18 & 5.21\% & \textbf{\textcolor{blue}{1.13}} & \textbf{\textcolor{blue}{5.04\%}} \\
V2 & 1.14 & 7.29\% & 1.01 & 6.31\% & 0.96 & 6.11\% & \textbf{\textcolor{blue}{0.96}} & \textbf{\textcolor{blue}{5.90\%}} \\
V3 & 4.77 & 17.35\%  & 4.76 & 17.33\% & 4.71 & 17.22\% &\textbf{\textcolor{blue}{4.65}} & \textbf{\textcolor{blue}{17.06\%}}\\
V4 & \textbf{\textcolor{blue}{2.82}} & \textbf{\textcolor{blue}{11.68\%}} & 2.88  & 11.84\% & 2.88  & 11.84\% & 2.88 & 11.84\% \\
V5 & 8.53 & 24.79\% & 8.10 & 23.48\% & 8.04 & 23.30\% & \textbf{\textcolor{blue}{8.03}} & \textbf{\textcolor{blue}{23.20\%}} \\
V6 & 10.54 & 38.61\%  & 10.10 & 37.30\% & 9.70 & 35.80\% &\textbf{\textcolor{blue}{9.25}} & \textbf{\textcolor{blue}{33.69\%}}\\
V7 & 2.97 & 9.57\% & \textbf{\textcolor{blue}{2.84}} & \textbf{\textcolor{blue}{9.31\%}} & 2.90 & 9.48\% & 2.99 & 9.73\% \\
V8 & 0.49 & 9.99\% & 0.28  & 5.13\% & 0.21  & 3.86\% & \textbf{\textcolor{blue}{0.11}} & \textbf{\textcolor{blue}{2.20\%}}\\
\hline
\end{tabular}
\end{table}

Compared with the curve of RO method using only one frame to predict the number of people which leads to an obvious oscillation, the QP method can predict more precisely by smoothing out the oscillations.
The proposed method is
highly effective to reduce the errors because of the nontrivial
network flow constraints. Otherwise, there will no improvements that can be made.
Such as in Fig.\ref{fig-res}(d), there is no remarkable promotion
between the regression-only method and the QP method. One of the reasons
is that there is only one group in most frames of this video, and
the only group connects either to vertex $S$ or to vertex $T$, so that no nontrivial network flow constraints are formed in the
network. Another reason is that the foreground
segmentation algorithm of inter-frame difference used in the paper is not robust enough, therefore, network flow constraints fail to improve the performance of video
V4.

The second group of tests is carried out on UCSD and Fudan datasets.  Since the foreground image of each frame is provided by the authors in
their datasets, our algorithm simply loads their foreground images in foreground extraction phase. For UCSD datasets, frames 601-1400 are used for training and the rest for testing. For Fudan datasets, frames 1-300 are used for training and the rest for testing. %Since the foreground image of each frame is given in these two datasets,  our algorithm simply load foreground images in foreground extraction phase.
Table \ref{tab-UCSDperformance} demonstrates that QPL$m$ methods can always reduce the errors of the two datasets.

\begin{table}[h]
\caption{Performance of RO method and  QPL$m$ on UCSD and fudan datasets.}\label{tab-UCSDperformance}
\centering
\begin{tabular}{ccccc}
\hline
\multirow{2}{*}{Method} &
\multicolumn{2}{c}{UCSD} &
\multicolumn{2}{c}{Fudan}  \\
\cline{2-3}\cline{4-5}
  & MAE & MRE & MAE & MRE  \\
\hline
RO & 2.41 &   10.02\% & 1.00 & 15.11\% \\

QPL7 & 2.33 & 9.58\%  & 0.93 & 13.72\%  \\

QPL15 & 2.27 & 9.22\% & 0.93 & 13.16\% \\

QPL23 & \textbf{\textcolor{blue}{2.22}} & \textbf{\textcolor{blue}{8.95\%}} & \textbf{\textcolor{blue}{0.93}} & \textbf{\textcolor{blue}{12.80\%}} \\
\hline
\end{tabular}
\end{table}

\begin{figure}[h]
\begin{minipage}[b]{0.5\linewidth}
  \centering
  \centerline{\includegraphics[width=8cm]{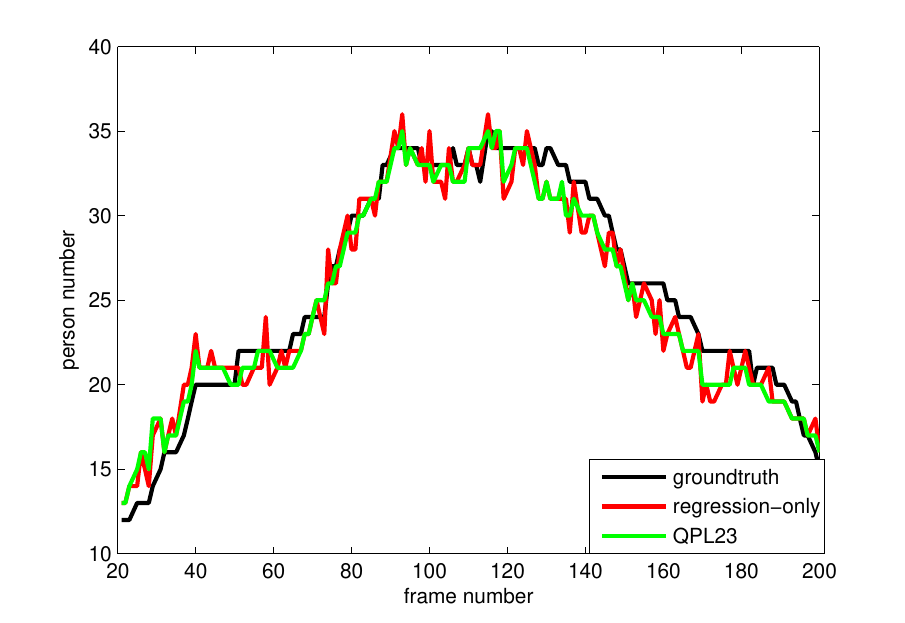}}
%  \vspace{2.0cm}
  \centerline{(a) V1}\medskip
\end{minipage}
\hfill
\begin{minipage}[b]{0.5\linewidth}
  \centering
  \centerline{\includegraphics[width=8cm]{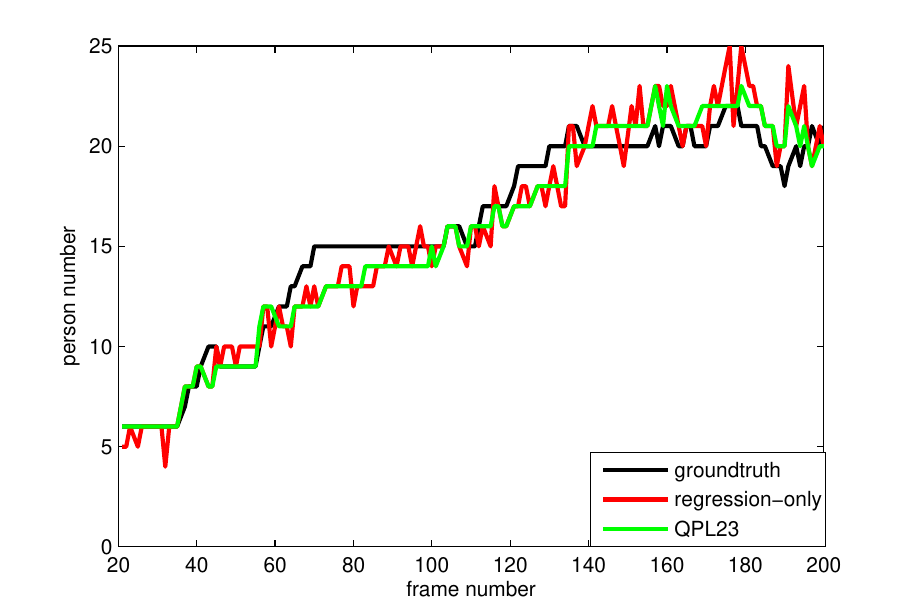}}
%  \vspace{2.0cm}
  \centerline{(b) V2}\medskip
\end{minipage}
\vspace{2.0cm}
\begin{minipage}[b]{0.5\linewidth}
  \centering
  \centerline{\includegraphics[width=8cm]{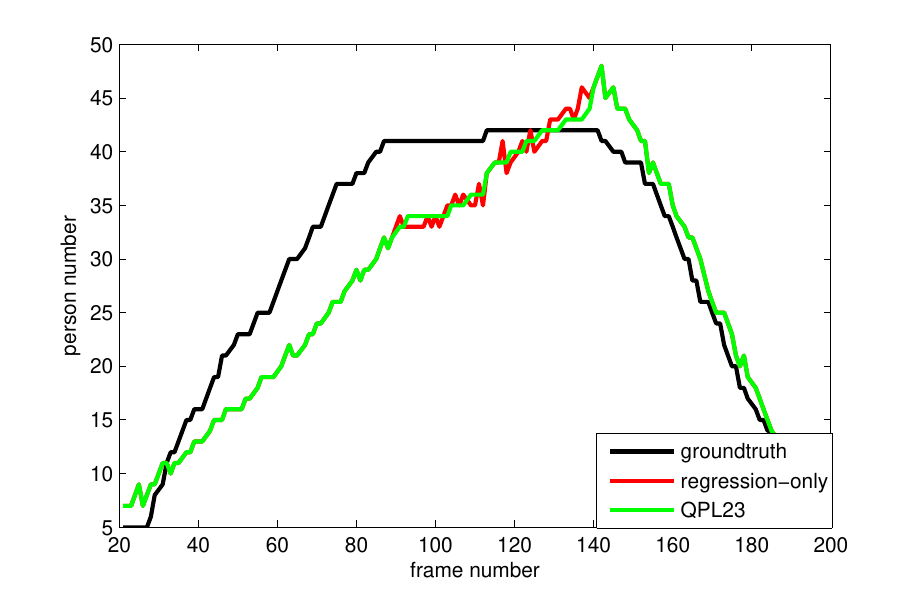}}
%  \vspace{2.0cm}
  \centerline{(c) V3}\medskip
\end{minipage}
\hfill
\begin{minipage}[b]{0.5\linewidth}
  \centering
  \centerline{\includegraphics[width=8cm]{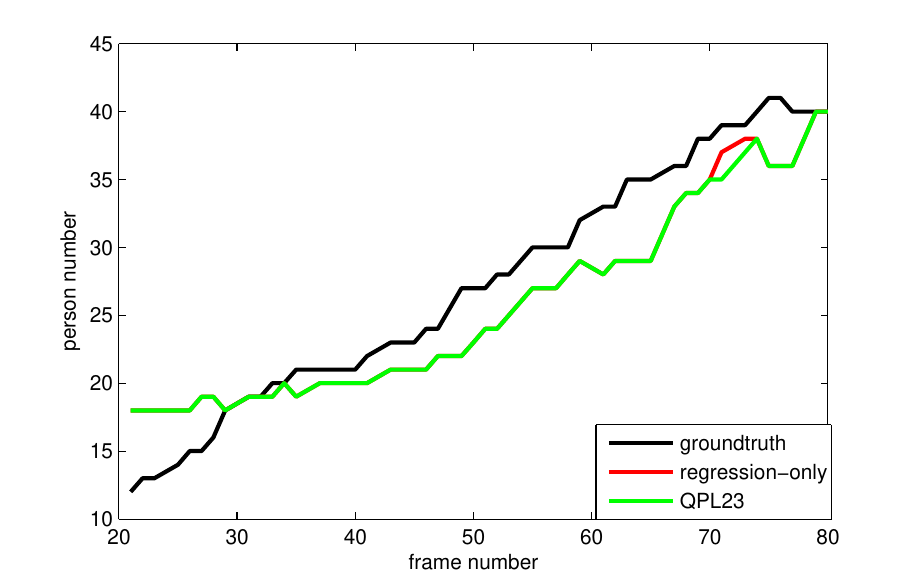}}
%  \vspace{2.0cm}
  \centerline{(d) V4}\medskip
\end{minipage}
\caption{Curves of count number estimated by different methods and the ground truth. x-axis presents frame number.}
\label{fig-res}
\end{figure}
\subsection{Evaluation of Processing Time}

 The datasets with the smallest and largest images
were selected for comparison: the UCSD dataset has a resolution of 236 * 158 pixels,
whereas the PETS 2009 dataset has a resolution of 768 * 576. The processing time of QPL$m$ methods is reported in Table \ref{tab-runtime}, which is obtained using a laptop with an Intel(R) Core(TM)i5-3317U CPU @1.70GHz and uses Visual Studio 2010 with Opencv2.4.8 on Windows 7. As shown in Table \ref{tab-runtime}, the  processing time of QPL$m$ methods  may increase significantly with the addition of new network layers. The time complexity of network flow equations mostly depends on the total groups in all layers in the constructed network. If there is only one group in most frames, the processing time of QPL$m$ methods has almost no noticeable changes with the increasing of the number of network layers, like in the video V3. Table \ref{tab-processing} presents the comparison of processing speed using different methods.

\begin{table}[h]
\caption{Processing time  per frame (ms) of the proposed algorithm with different datasets.}\label{tab-runtime}
\centering
\begin{tabular}{ccccccc}
\hline
%\multirow{2}{*}{Method} &
%\multicolumn{2}{c}{UCSD} &
%\multicolumn{2}{c}{Fudan}  \\
%\cline{2-3}\cline{4-5}
%  & MAE & MRE & MAE & MRE  \\
dataset & V1& V2 &V3 &V4 & UCSD & Fudan \\
\hline
resolution & 768*576 &  768*576 & 768*576 & 768*576 &238*158 & 320*240\\
RO  & 152 & 145 & 149 & 146  &25  & 46 \\
%QPL5  & 154 & 146 & 149 & 146  &26  & 47 \\
QPL7  & 155 & 146 & 149 & 147  &28  & 49  \\
%QPL9  & 156 & 147 & 149 & 147  &34  & 51  \\
%QPL11 & 160 & 149 & 148 & 146  &38  & 54 \\
%QPL13 & 164 & 152 & 148 & 147  &49  & 56 \\
QPL15 & 173 & 156 & 149 & 148  &65  & 63 \\
%QPL17 & 184 & 163 & 151 & 148  &90  & 75 \\
%QPL19 & 199 & 172 & 150 & 149  &129 & 86\\
%QPL21 & 220 & 183 & 149 & 150  &180 & 106 \\
QPL23 & 255 & 197 & 149 & 151  &254 & 134 \\
\hline
\end{tabular}
\end{table}

\begin{table}[h]
\caption{Comparison of processing speed(fps) with different algorithms}\label{tab-processing}
\centering
\begin{tabular}{cccc}
\hline
Method & PETS2009 &
UCSD &
Fudan  \\
\hline
Ryan\cite{RDD} &4.5 &32.3 &--  \\
Al-Zaydi\cite{AZZQH}    &\textbf{\color{blue}14.88} & -- &--\\
QPL3 &6.9 &\textbf{\color{blue}40.00}& \textbf{\color{blue}21.74}\\
QPL23 &5.88 & 3.94& 7.46\\
\hline
\end{tabular}
\end{table}

\subsection{Comparison with other methods}

Table \ref{tab-performance-PETS2009} presents the comparison between the counting accuracy of the proposed method and that of Conte's\cite{CDF2} and Zini's\cite{ZLN} method on PETS2009 datasets. Here, the training set is constructed by manually collecting 30-40 frames from each video, which is the same as that in Conte's method. It is worth noting that when compared with Conte's results the QP method has a significant performance improvements, except video V4 and video V5. In these two videos, there is a rapid change of the density of people when they turn right in the scene. Our designed feature, too simple to adapt well to these rapid changes, may to some extent affect initial estimations of people number. Despite of the improvements on the results the QP method can make, there are still gaps between our results and Conte's for the two videos.
\begin{table}[htp]
\caption{Performance of our method and other methods on PETS2009 datasets.}\label{tab-performance-PETS2009}
\centering
\begin{tabular}{ccccccc}
\hline
\multirow{2}{*}{Method} &
\multicolumn{2}{c}{Conte\cite{CDF2}} &
\multicolumn{2}{c}{Zini\cite{ZLN}} & \multicolumn{2}{c}{QPL23}  \\
\cline{2-3}\cline{4-5}\cline{6-7}
   & MAE & MRE& MAE & MRE& MAE & MRE \\
\hline
V1  & 1.36 & 6.80\%&1.8 &- & \textbf{\textcolor{blue}{1.13}} & \textbf{\textcolor{blue}{5.04\%}} \\
V2  & 2.55 & 16.30\% & 1.72 &- & \textbf{\textcolor{blue}{0.96}} & \textbf{\textcolor{blue}{5.90\%}} \\
V3  & 5.40 & 20.80\% &\textbf{\textcolor{blue}{2.01}} &-  & 4.65 & 17.06\% \\
V4   & 2.81 & 15.10\%  &\textbf{\textcolor{blue}{2.0}} &- & 2.88 & 11.84\%\\
V5  & \textbf{\textcolor{blue}{4.45}} & \textbf{\textcolor{blue}{15.10\%}}  &- &- & 8.03 & 23.20\% \\
V6 & 12.17 & 30.70\% &- &- & \textbf{\textcolor{blue}{9.25}} & \textbf{\textcolor{blue}{33.69\%}} \\
V7  & 7.55 & 23.60\%  & -&- & \textbf{\textcolor{blue}{2.99}} & \textbf{\textcolor{blue}{9.73\%}}\\
V8  & 1.64 & 35.2\%  & -&-& \textbf{\textcolor{blue}{0.11}}& \textbf{\textcolor{blue}{2.20\%}}\\
\hline
\end{tabular}
\end{table}

Table \ref{tab-performance-PETS20092} demonstrates the comparison between the performance of the proposed method and that of Zhang's\cite{ZXH}  method on PETS2009 datasets. The training sets are the same as that in Zhang's\cite{ZXH} method as reported in Table \ref{tab-performance-PETS20092}. The results suggest that the regression model with network flow constraints can always obtain lower MRE. That is, we can picture that there are no sharp changes on the curve of people counting predicted by the proposed method.

\begin{table}[htp]
\caption{Performance of our method and Zhang's methods on PETS2009 datasets.}\label{tab-performance-PETS20092}
\centering
\begin{tabular}{ccccc}
\hline
\multirow{2}{*}{Method} &
\multicolumn{2}{c}{Zhang\cite{ZXH}} &
 \multicolumn{2}{c}{QPL23}\\
\cline{2-3}\cline{4-5}
   & MAE & MRE& MAE & MRE\\
\hline
S1.L1.13-57(1)  & training & - & training & - \\
S1.L1.13-59(1)  & \textbf{\textcolor{blue}{2.15}} & 13.86\% &2.22 &\textbf{\textcolor{blue}{13.60\%}}\\
S1.L2.14-06(1)  & 9.89 & 34.87\% &\textbf{\textcolor{blue}{8.65}} &\textbf{\textcolor{blue}{27.12\%}} \\
S1.L1.13-57(2)  & training & - & training & - \\
S1.L2.14-06(2)  & \textbf{\textcolor{blue}{9.98}} &62.21\% &14.58 &\textbf{\textcolor{blue}{36.58\%}}\\

\hline
\end{tabular}
\end{table}

Finally, we compare our proposed method with previous studies in the literature on the UCSD and Fudan datasets in Table \ref{tab-performance-UCSD}.
As shown in Table \ref{tab-performance-UCSD}, the proposed method can get lower MRE than Ryan's algorithm. However, QP method performs less than successfully on UCSD dataset. Since we simply load foreground images provided by authors of UCSD dataset in foreground extraction phase and the provided foreground area is larger than the real area, because the foreground area includes the moving people and some background pixels. The group feature, which uses all pixels in each group as key part, interferes with the accurate estimation of the number of people in this method.

\begin{table}[htp]
\caption{Performance of the proposed method and other methods on UCSD and Fudan datasets.}\label{tab-performance-UCSD}
\centering
\begin{tabular}{ccccccccc}
\hline
\multirow{2}{*}{Method} &
\multicolumn{2}{c}{Ryan\cite{RDD}} &
\multicolumn{2}{c}{Al-Zaydi\cite{AZZQH}} &
\multicolumn{2}{c}{Conte\cite{CDF2}} &
\multicolumn{2}{c}{QPL23}  \\
\cline{2-3}\cline{4-5}\cline{6-7}\cline{8-9}
   & MAE & MRE& MAE & MRE& MAE & MRE &  MAE & MRE \\
\hline
UCSD   &\textbf{\textcolor{blue}{1.46}} &6.23\%   &1.63 & \textbf{\textcolor{blue}{4.32\%}}  & 3.26 & 10.88\% &2.22& 8.95\%\\
Fudan  &\textbf{\textcolor{blue}{0.92}} &15.51\%  &-&- &-&- &0.93 &\textbf{\textcolor{blue}{12.80\%}}\\
\hline
\end{tabular}
\end{table}

\section{Conclusion}\label{sec-conclusion}

This paper introduced the network flow constraints to crowd counting for the first time. An integer quadratic programming model is used to improve the prediction of regression methods. To put it simple, the integer variables are firstly relaxed into real ones. Then the integer quadratic programming models can be solved with linear equations. And the experimental results show that the proposed algorithm can enhance the accuracy of the RO methods significantly in a great majority of videos. When compared with other methods, the QPL$m$ algorithm shows the obvious improvement and lower MRE in most videos. In addition, the time complexity can be controlled within the acceptable range by adapting the number of network layers. In the future, the more precise foreground segmentation algorithms and more complex group features will be explored to improve the performance.
And the original integer programming problem should be considered directly, since the relaxation of integer variables will lead to only approximate solutions.

\section{Acknowledgment}\label{sec-thanks}
The authors would like to express their gratitude to the anonymous referees for their kind suggestions and comments on the original manuscript, which resulted in this vision. This work is supported by Major Program of National Natural Science Foundation of China (grant 71533001) and the Key Projects in the National Science \& Technology Pillar Program during the Twelfth Five-Year Plan Period(grant 2013BAK02B06).
%This research was support by Key Project in the National Science \& Technology Pillar Program During the Twelfth Five-year Plan Period(2013BAK02B06-03) and National Natural Science Foundation of China .


\end{document}